\documentclass{ecai}
\usepackage{graphicx}
\usepackage{caption}
\usepackage{latexsym}

\usepackage{xcolor}
\usepackage{xspace}
\usepackage{dsfont} 

\usepackage[ruled,vlined]{algorithm2e}
\usepackage[noend,compatible]{algpseudocode}

\graphicspath{{images/}{../images/}}

\usepackage{amsmath}
\usepackage{amsfonts}
\usepackage{amssymb}
\newtheorem{lemma}[theorem]{Lemma}

\usepackage{tabularray}

\newcommand{\1}[1]{\mathds{1}\lbrace #1 \rbrace\xspace}
\newcommand{\bbE}{\mathop{\mathbb{E}}\xspace}

\renewcommand {\cal}[1]{\ensuremath{{\mathcal{#1}}}\xspace}
\renewcommand {\bold}[1]{\ensuremath{{\mathbf{#1}}}\xspace}

\newcommand\norm[1]{\left\lVert#1\right\rVert\xspace}

\newcommand\zeroOne{0\mbox -1}
\newcommand{\zeroOneloss}{\ell^{\zeroOne}\xspace}

\begin{document}

\begin{frontmatter}

\title{Adversarial attacks for mixtures of classifiers}

\author[A]{\fnms{Lucas}~\snm{Gnecco Heredia}\orcid{0000-0002-1561-2080}\thanks{Corresponding Author. Email: lucas.gnecco-heredia@lamsade.dauphine.fr.}}
\author[A]{\fnms{Benjamin}~\snm{Negrevergne}\orcid{0000-0002-7074-8167}}
\author[A]{\fnms{Yann}~\snm{Chevaleyre}\orcid{0000-0002-6609-5562 }} 

\address[A]{LAMSADE, CNRS, Université Paris Dauphine - PSL}


\begin{abstract}
Mixtures of classifiers (a.k.a. randomized ensembles) have been proposed as a way to improve robustness against adversarial attacks. However, it has been shown that existing attacks are not well suited for this kind of classifiers. In this paper, we discuss the problem of attacking a mixture in a principled way and introduce two desirable properties of attacks based on a geometrical analysis of the problem (effectiveness and maximality). We then show that existing attacks do not meet both of these properties. Finally, we introduce a new attack called {\em lattice climber} attack with theoretical guarantees on the binary linear setting, and we demonstrate its performance by conducting experiments on synthetic and real datasets.
\end{abstract}

\end{frontmatter}

\section{Introduction}

Deep neural networks have been shown to be vulnerable to adversarial attacks \cite{szegedy2013adv}, i.e. small perturbations that, although imperceptible to humans, manage to drastically change the predictions of the model. This observation has led to numerous efforts to understand this phenomenon \cite{fawzi2018analysis, bubeck2019adversarial} and started a series of publications introducing various techniques to train robust models \cite{goodfellow2014explaining, madry2017towards} as well as new algorithms to attack them \cite{carlini2017towards, tramer2020adaptive, croce2020autoattack}.

One possible way to build a robust classifier is to train a diverse collection of classifiers and to select one at random at inference time to make a prediction. The resulting randomized classifier is called a {\em randomized mixture} (or a randomized ensemble \cite{dbouk2022arc,dbouk2023robustness}), and is provably more robust than any of the classifiers in the collection when faced with a regularized adversary~\cite{pinot2020randomization}. 

This first theoretical result was later followed by~\cite{meunier2021mixed} that showed the existence of a Nash equilibrium when both players (the model and the attacker) are allowed to use mixed (i.e. random) strategies, giving additional theoretical arguments in favor of using randomized mixtures as a way to develop robust models with strong theoretical guarantees.

Despite a number of promising theoretical results in favor of randomized mixtures, there has been comparatively fewer publications on the problem of developing attacks that are effective in practice against them. In the past, the lack of efficient and specialized attacks has commonly undermined the reliability of the empirical evaluation of various defense mechanisms~\cite{tramer2020adaptive}. Indeed, the authors of \cite{dbouk2022arc} have shown that the attack used to evaluate the robustness of mixtures was not adequate, leading to a large overestimation of the empirical robustness of the model proposed in \cite{pinot2020randomization}.

In this work, we take a principled approach towards understanding adversarial attacks for mixtures of classifiers using a set theoretic perspective, where the sets are the {\em vulnerability regions} of each classifier of the mixture. We then show that the problem of attacking a mixture can be seen as the problem of exploring a lattice. Using this perspective, we identify a series of desirable properties, and devise a new attack that satisfies these properties and is efficient in practice. 

More specifically, contributions are the following:

\begin{itemize}
    \item We model the problem of attacking a mixture with an intuitive partially ordered set, more specifically a lower semi-lattice, which allows us to better characterize existing attacks and to identify the optimal behavior an attacker should have.
    \item With this framework in mind, we propose an attack algorithm that has strong guarantees in the binary linear setting compared to existing attacks. We then generalize our proposed attack to multi-class differentiable classifiers.

\end{itemize}

Our code will be made available upon acceptance of this paper. 

\section{Preliminaries}

\paragraph{Notations. } In this work, we denote $\Delta^k$ the probability simplex in $\mathbb{R}^k$, and $\{e_1, \dots e_k\}$ the vertices of $\Delta^k$, where $e_j$ represents the \textit{one hot} encoding of $j$ and is the vector whose components are all zero, except the component $j$ that equals one. For a probability vector $p~\in~\Delta^k$, we denote $Cat(p)$ the categorical distribution over $k$ elements. For a predicate $C$, we denote $\1{C}$ the function that equals 1 if the predicate $C$ is true, 0 otherwise. For an integer $m$, we use the notation $\left[m\right]=\{1,\cdots ,m\}$.

\paragraph{Problem setting.}
Given a $d\mbox -$dimensional input space $\mathbb R^d$ and a set of $k$ class labels, a deterministic classifier $h : \mathbb R^d \to \{e_1, \dots, e_k\}$ is a function that maps each input point $x$ to a predicted class represented using its one hot encoding\footnote{For convenience, in the binary classification setting, we may also represent a classifier as a function mapping $\mathbb{R}^d$ to either $\{0, 1\}$ or $\{-1, 1\}$.}. We can compute the $0\mbox -1$ loss of $h$ at $x$, denoted $\zeroOneloss(h, x, y)$ as follows:
\begin{equation}\label{eq:zero_one_loss}
    \zeroOneloss(h, x, y) = \1{h(x) \neq e_y}
\end{equation}

In this paper, we consider classifiers of the form $h~:~\mathbb R^d~\rightarrow~\Delta^k$, where the vector $h(x)$ is interpreted as a probability distribution over the $k$ classes. This allows the more general definition of \textit{randomized} or \textit{probabilistic} classifiers, for which one inference step would require sampling a class from $Cat(h(x))$ (See \cite[Definition 1]{pinot2022robustness}). The $0\mbox -1$ loss needs to be generalized to include the expectation over the class distribution given by probabilistic classifier $h$ \cite[Equation 6]{pinot2022robustness}, as follows:
\begin{equation}\label{eq:expected_zero_one_loss}
    \zeroOneloss(h, x, y) = \bbE_{\hat{y} \sim h(x)} \1{e_{\hat{y}} \neq e_y}
\end{equation}

\paragraph{Mixture of classifiers.}
Given a set ${\bf h} = \{h_1, \cdots, h_m\}$ of $m$ deterministic classifiers and a discrete probability distribution $\bf q$ over $\left[ m \right]$ that assigns a probability to each classifier $h_i \in {\bf h}$, we can build a {\em mixture of classifiers}  ${\bf m_q^h}$ that assigns a class label to $x$ by first sampling an index $i$ from the distribution $Cat(q)$, and then returning the label $h_i(x)$. To obtain the probability distribution in $\Delta^k$ corresponding to a particular example $x$, we have to compute the following weighted sum: 
$${\bf m_q^h}(x) = \sum_{i=1}^{m} q_i \cdot h_i(x)$$

Moreover, the $\zeroOneloss$ which we defined earlier for the general case of probabilistic classifiers can now be re-written for the specific case of mixtures as follows:
\begin{equation}\label{eq:zero_one_mixture}
\zeroOneloss({\bf m_q^h}, x, y) = \sum_{i=1}^{m} q_i \cdot \1{h_i(x) \neq e_y} 
\end{equation}

\paragraph{Adversarial attacks on classifiers and mixtures.}

Given an input point $x \in \mathbb R^d$ and its true label $y$, attacking a classifier $h$ (deterministic or probabilistic) consists in discovering a norm bounded perturbation $\delta \in \mathbb R^d$ (with $\norm{\delta} \le \epsilon$) that increases $\zeroOneloss(h, x + \delta, y)$. Various norms can be used to measure the magnitude of the perturbation $\delta$, the most common being $\ell_p$ norms with $p = 2$ or $p = \infty$. In the rest of this paper, we assume $p=2$, but the results also hold for $p = \infty$.  We denote $B^{\epsilon}(x)$ the corresponding ball centered in $x$ with radius $\epsilon$ \textit{i.e.} $B^\epsilon(x) = \{ x + \delta  \mid \norm{\delta} \le \epsilon \}$ and we call $x+\delta \in B^\epsilon(x)$ an {\em adversarial example} of $x$.

Formally speaking, the adversarial $0 \mbox-1$ loss denoted $\zeroOneloss_{\epsilon}$ is simply the $0 \mbox-1$ loss under attack of an optimal adversary and is defined as follows: 
\begin{equation}
    \zeroOneloss_{\epsilon}(h, x, y)\quad= \sup\limits_{\norm{\delta} \le \epsilon} \zeroOneloss(h, x + \delta, y)
\end{equation}
For mixtures of classifiers, the problem of maximizing the $0 \mbox-1$ loss around $x$ is equivalent to finding a perturbation that fools the coalition of base classifiers with the highest weight:
\begin{equation} \label{eq:zero_one_mixture_adversarial}
  \zeroOneloss_{\epsilon}({\bf m_q^h}, x, y) = \sup\limits_{\norm{\delta} \le \epsilon} \sum_{i=1}^{m} q_i \cdot \1{h_i(x + \delta) \neq e_y}
\end{equation}

\section{Attacking a mixture of classifier} \label{sec:geometry}

In the following section, we provide a geometrical analysis of the problem of attacking a mixture, which shed some light on the optimal behavior to be adopted by the attacker as well as the limits of existing attacks. We start by introducing the concept of vulnerability region, which is at the core of this analysis.

 \subsection{Vulnerability regions}
 
Let $x$ be a point with correct class $y$, and fix $\epsilon > 0$ some budget for the attacker. For a single deterministic classifier $h$, the vulnerability region of $h$, denoted $V(h)$, is the set that contains all the adversarial examples of $h$ around $x$ for the fixed budget $\epsilon$, i.e. $V(h) = \{ x' \in B^\epsilon(x) \mid h(x') \neq e_y\}$.

When considering a set of classifiers ${\bf h}$, we define its vulnerability region as the intersection of the individual vulnerability regions, \textit{i.e.} $V(\bold h) = \bigcap_{h \in \bold h} V(h)$.
Note that if $V(\bold h) = \emptyset$, then there is no adversarial example capable of fooling all classifiers simultaneously, and a mixture built from $\bold h$ is guaranteed to show at least some level of robustness at $x$.
To simplify notation, when considering subsets of classifiers from a set $\bold h$ of $m$ classifiers, we will also use their indices $\cal I \subseteq \left[ m \right]$ and refer to their vulnerability region as $V(\cal I)$. For the sake of coherence, by convention we set $V(\emptyset) = B^{\epsilon}(x)$.

For example, in Figure\ref{two_classifiers_regions}.(c), $V(\{1\})$ corresponds to the orange region, $V(\{2\})$ to the blue region and $V(\{1, 2\}) = \emptyset$, whereas in Figure\ref{two_classifiers_regions}.(d), $V(\{1, 2\})$ is the green region.

Let us consider a two classifier mixture ${\bf m_q^h}$, \textit{i.e.} $\bold h = (h_1, h_2)$. 
Thanks to the relative simplicity of this setting, we can conduct a case study for each one of the four possible configurations (i.e. different spacial arrangement of the two classifiers). The four possible configurations are illustrated in Figure~\ref{two_classifiers_regions} from the most convenient for the defender (configuration $(a)$ on the left) to the most convenient for the attacker (configuration $(d)$ on the right).

\newcommand\figscale{0.2}
\begin{figure*}[ht]
    \centering
    \captionsetup{width=.25\linewidth,labelformat=empty}
    \minipage{0.25\textwidth}
    \centering
      \includegraphics[scale=\figscale]{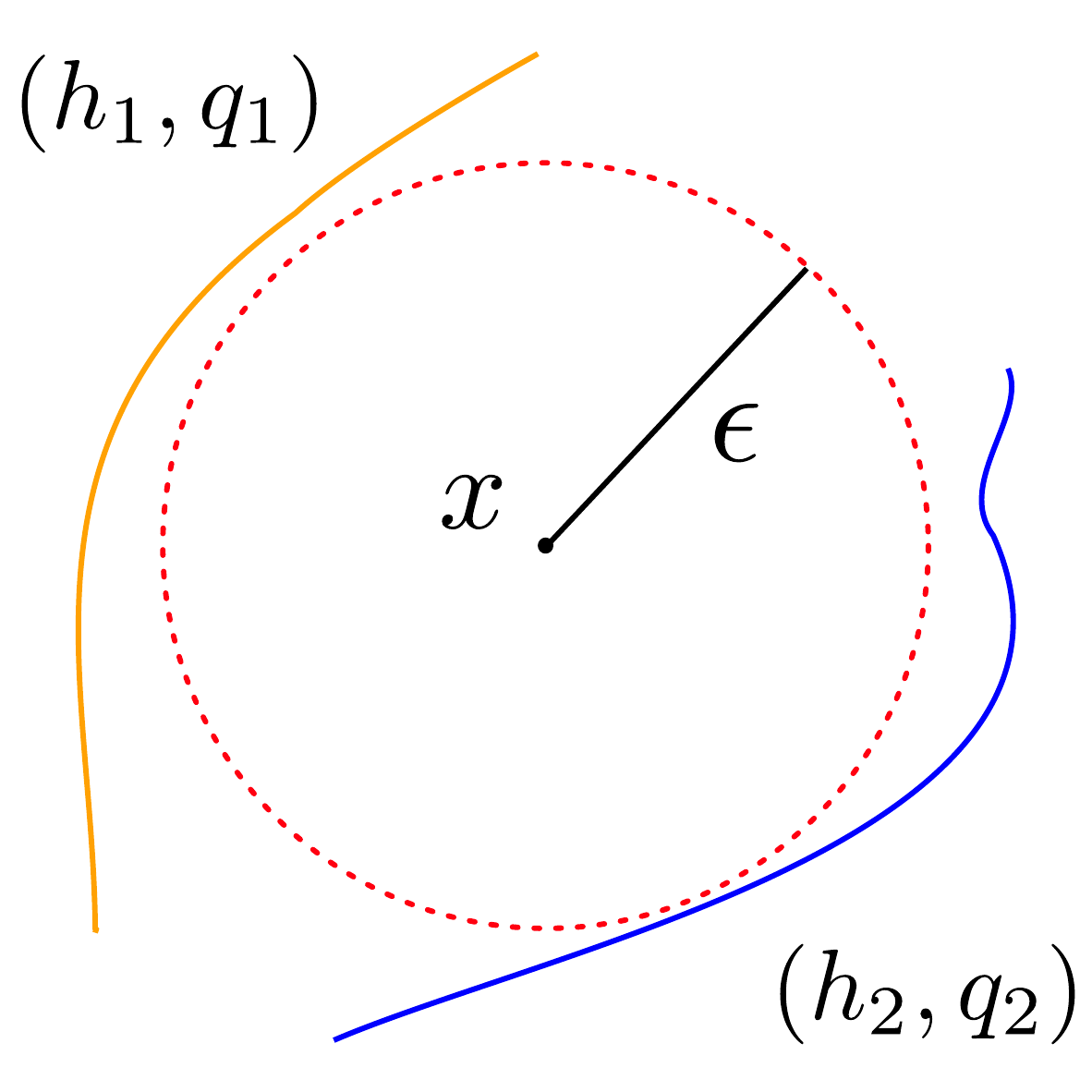}
      \caption{a) Both robust}
      \label{two_classifiers_regions_rob}
    \endminipage\hfill
    \minipage{0.25\textwidth}%
    \centering
      \includegraphics[scale=\figscale]{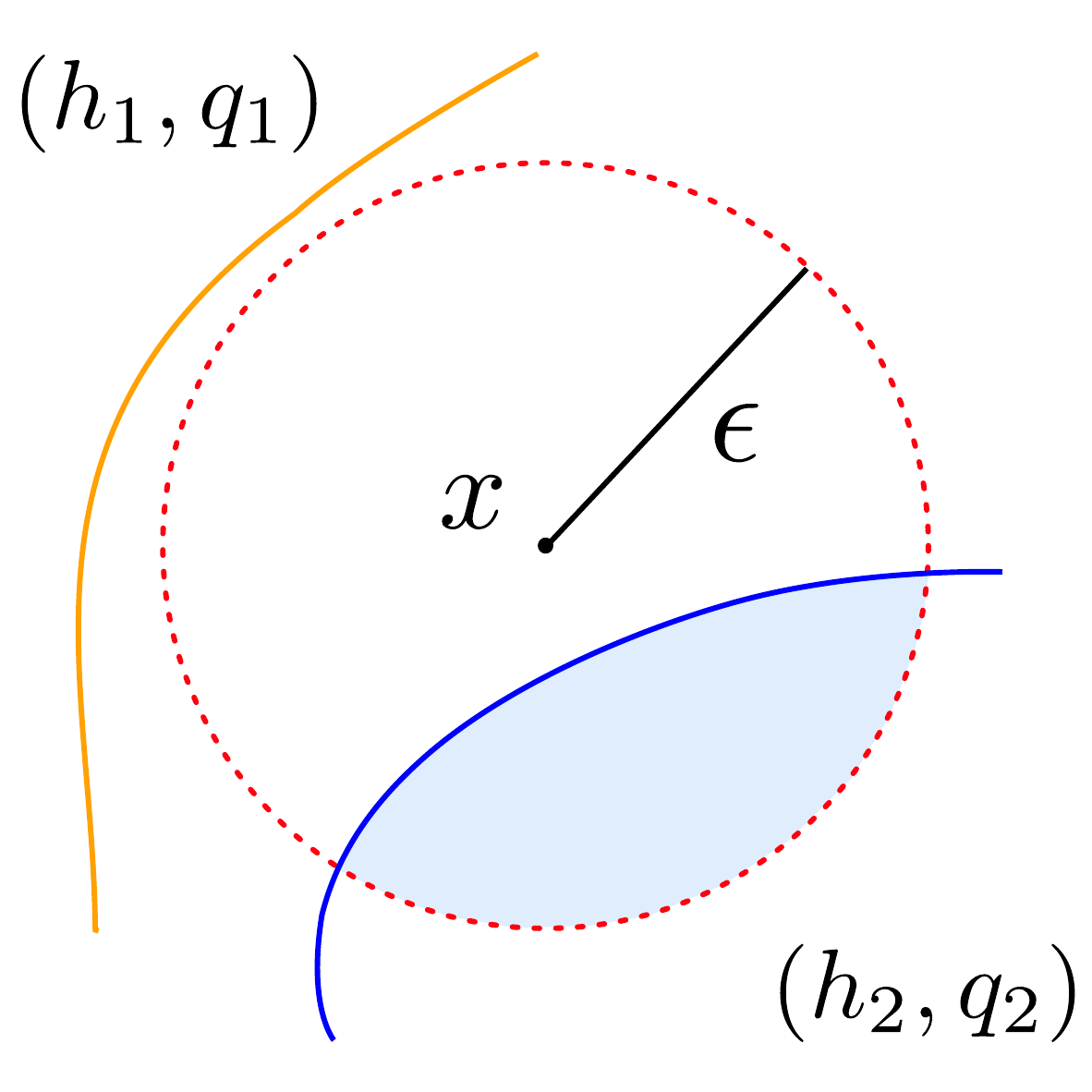}
      \caption{b) Only one robust}
      \label{two_classifiers_regions_onerob}
    \endminipage
    \minipage{0.25\textwidth}
    \centering
      \includegraphics[scale=\figscale]{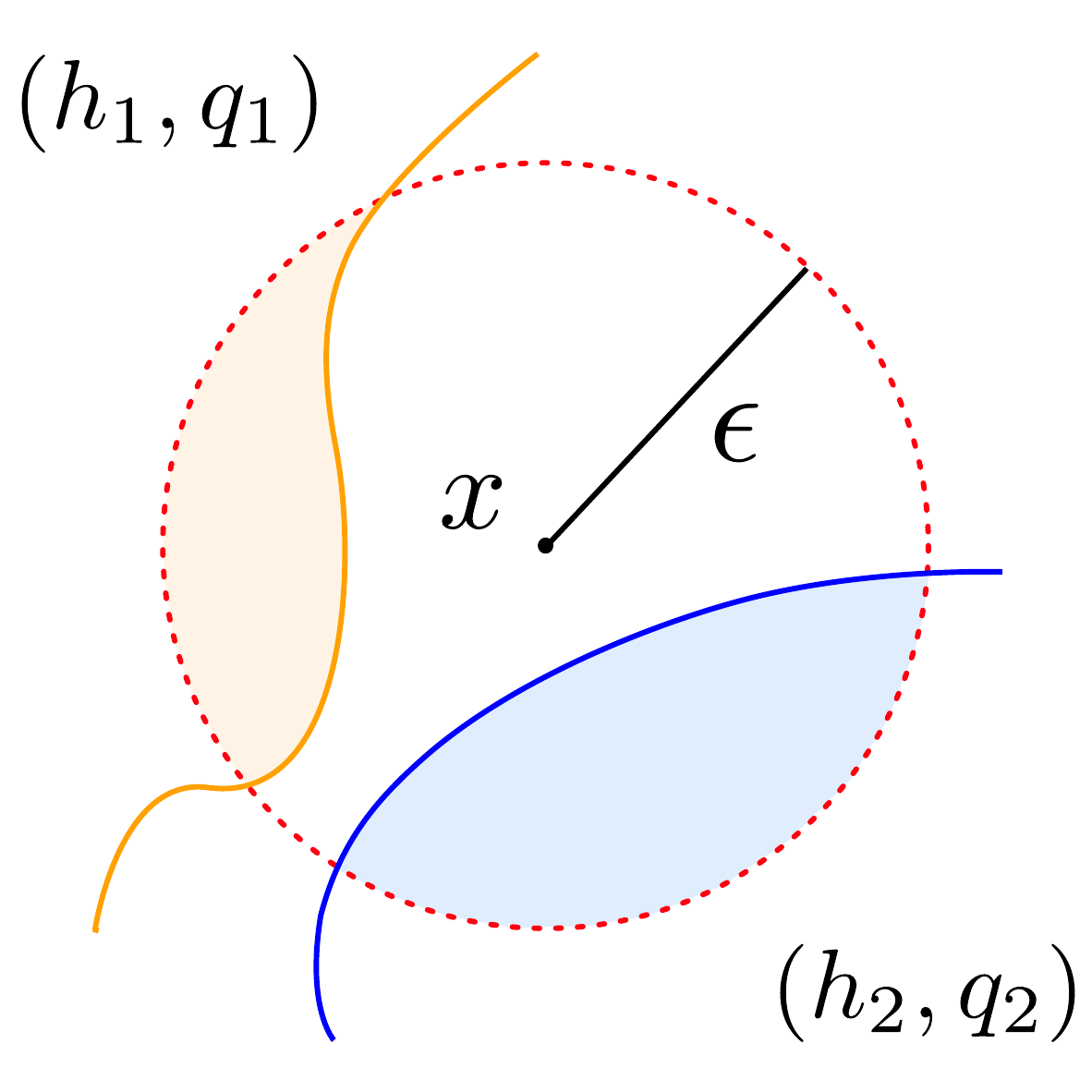}
      \caption{c) No intersection}
      \label{two_classifiers_regions_mp}
    \endminipage\hfill
    \minipage{0.25\textwidth}%
    \centering
      \includegraphics[scale=\figscale]{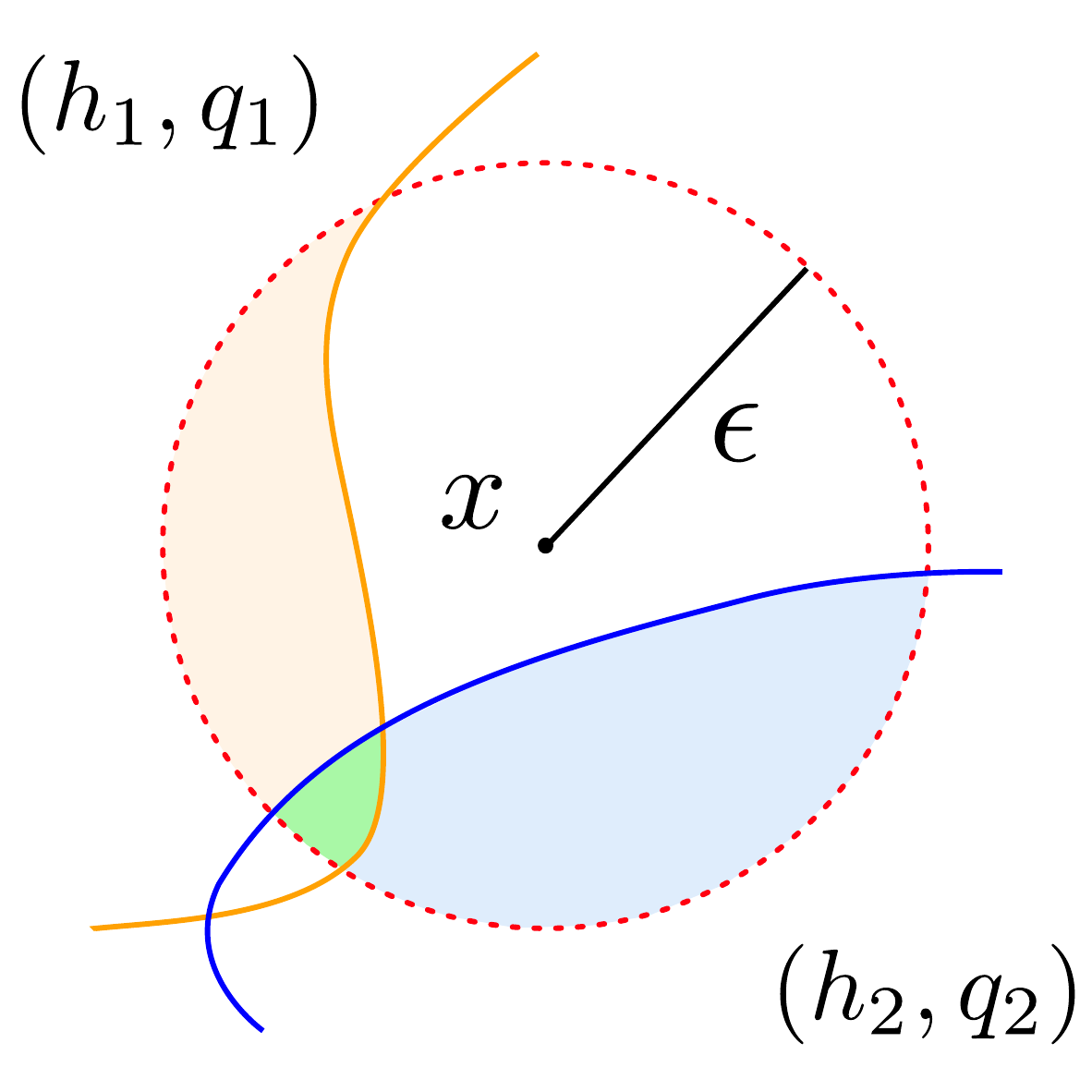}
      \caption{d) Intersection}
      \label{two_classifiers_regions_intersection}
    \endminipage
    \addtocounter{figure}{-1}
    \addtocounter{figure}{-1}
    \addtocounter{figure}{-1}
    \addtocounter{figure}{-1}
    \captionsetup{width=1.0\linewidth, labelformat=original}
    \caption{Four possible configurations for a mixture with two classifiers $h_1$ and $h_2$ assuming $x$ is correctly classified by both. a) Both $h_1$ and $h_2$ are non-vulnerable, the mixture is robust. b) Only one classifier is vulnerable c) Both $h_1$ and $h_2$ are vulnerable, but they can not be attacked simultaneously d) Both $h_1$ and $h_2$ are vulnerable, and they can be attacked on the same region. Best viewed in color.}
    \label{two_classifiers_regions}
\end{figure*}

Configuration $(a)$ is analogous to the traditional notion of robustness because no perturbation can increase the $\zeroOne$ loss, \textit{i.e.} $\zeroOneloss_{\epsilon}({\bf m_q^h}, x, y) = 0$. In configuration $(b)$ only $h_2$ is vulnerable, so an optimal attack would target $h_2$ to obtain $\zeroOneloss_{\epsilon}({\bf m_q^h}, x, y) = q_2$. In this case, the attack algorithm can ignore $h_1$ to craft the optimal perturbation.

For configuration $(c)$, in which both classifiers cannot be attacked at the same time, \textit{i.e.} $V(\bold h) = \emptyset$, the optimal attack is to target the classifier with the highest weight. As there does not exist a perturbation that can attack simultaneously both classifiers, targeting them both at the same time (for example with a gradient based optimization method) will most likely produce a perturbation that is non-adversarial for neither one of them. On the other hand, for configuration $(d)$, the optimal attack targets both classifiers at the same time. In this case, attacking any of the individual classifiers would increase the $\zeroOne$ loss to $q_1$ or $q_2$, but it is preferable to attack on the intersection $V(h_1) \cap V(h_2)$ to get a $\zeroOne$ loss of 1.

\subsection{Desirable properties of an attack for mixtures}

Our goal is to design an attack algorithm satisfying some desirable properties. In this section, we list three desirable properties, from the weakest to the strongest.

\paragraph{Effectiveness property.} In this work, we say an attack algorithm is \textit{effective} if for any point, it is able to generate an adversarial example increasing the $\zeroOne$ loss of the mixture when possible. In \cite{dbouk2022arc}, the authors criticize Adaptive Projected Gradient Descent (APGD), the attack used in \cite{pinot2020randomization} to evaluate the robustness of a mixture, because of its lack of effectiveness\footnote{In \cite{dbouk2022arc}, the authors named it \emph{consistency} instead of effectiveness. We find the latter more appropriate}. More precisely, they show that in situations like configuration $(c)$, APGD fails to find a successfull attack, because it tries to attack both classifiers at the same time. This motivates the authors to create \textit{Attacking Randomized ensembles of Classifiers} (ARC)\cite{dbouk2022arc}, an attack that is proven to be \textit{effective} in the binary linear classifier case.

Being effective solves the problem that APGD had in configuration $(c)$. However, effectiveness is not the only desirable property: in configuration $(d)$, attacking in the region $V(\bold h)$ is preferable than just guaranteeing an attack in some $V(\cal I),~\cal I \subseteq [m]$. When the number of classifiers in the mixture becomes large, having the guarantee that at least one classifier will be successfully attacked becomes weaker. This motivates the question: \textit{Can we design an algorithm with a stronger guarantee than effectiveness?}

\paragraph{Maximality.}
The idea that in configuration $(d)$, the region $V(\bold h)$ is preferable than any other vulnerability region can be formalized with the concept of \textit{maximal vulnerability region.} Given a set of classifiers $\bold h$ and point $(x, y)$, a vulnerability region defined by the classifiers $\lbrace h_i \rbrace_{i \in \cal I}$ indexed by $\cal I \subseteq \left[ m \right]$ is said to be maximal if it is non-empty ($V(\cal I) \neq \emptyset$) and
$$\forall j \in \left[ m \right] \setminus \cal I, \hspace{4px} V(\cal I \cup \{ j \} ) = \emptyset$$
We say an attack algorithm is \textit{maximal} if it guarantees that, in every case, the produced adversarial example belongs to a maximal vulnerability region. Note that in configuration $(c)$, effectiveness and maximality guarantees coincide. In configuration $(d)$, however, the maximality guarantee becomes stronger than effectiveness.

\paragraph{Optimality.}

The most desirable property of an algorithm attacking mixtures would be \emph{optimality}, which can be defined as the guarantee to find the adversarial attack achieving the highest possible 0-1 loss.
Unfortunately, no polynomial-time algorithm is guaranteed to achieve optimality, even when using linear classifiers, due to the following result:

\begin{theorem}(Hardness of attacking linear classifiers). Consider a binary classification setting. Given a labeled point
$(x,y)$, a set of $m$ linear classifiers $x\mapsto\mathds{1}\left\{ \theta_{i}^{\top}x+b_{i}\ge0\right\} $
where $(\theta_{i},b_{i}) \in\mathbb{R}^{d+1}$, a uniform mixture ${\bf m}$
composed of these linear classifiers, a noise budget $\epsilon>0$
and a value $\beta>0$, the problem of checking if there exists $\delta\in B^{\epsilon}(x)$ such $\ell_{\epsilon}^{0-1}\left({\bf m},x+\delta,y\right)\ge\beta$
is NP-hard.
\end{theorem}

Here is an intuition of the proof (full proof in Appendix \ref{app:proofs}): In the simplified setting of binary classification with linear classifiers, \textit{i.e.} each $h_i$ is a hyperplane, attacking an individual classifier $h_i$ can be formulated as satisfying a linear inequality, so solving equation \ref{eq:zero_one_mixture_adversarial} is highly related to the \textit{maximum feasible linear subsystem } \verb|(MaxFLS)| problem, which was proven to be NP-hard \cite{amaldi1995complexity}. In \cite[Theorem 2, Appendix E]{perdomosinger} the authors prove a weaker version of this result.

As a consequence, our algorithms will not aim at achieving optimality. Instead, we will focus on maximality, which will give us effectiveness for free.

\section{LCA attack} \label{our_attack}

To continue improving in the development of attacks against mixtures, we develop a new attack called Lattice Climber Attack (LCA) inspired on the maximality property. We will first present a simpler version of our attack that has maximality guarantees when faced with a mixture of binary linear classifiers. We will then extend it to multi-class differentiable classifiers.

To better understand LCA, we will develop in more depth the concept of preference between vulnerability regions that was introduced in Section \ref{sec:geometry}.

\paragraph{Order relation between vulnerability regions.}
Let ${\bf m_q^h}$ be a mixture. Let us consider two subsets of classifiers of $\bold h = \{ h_1, \cdots, h_m\}$ and represent them by the index sets $\cal I$, $\cal J \subseteq \left[ m \right]$. One can verify that if $\cal I \subseteq \cal J$ then $V(\cal I) \supseteq V(\cal J)$ and that additionally, if $V(\cal J) \neq \emptyset$, then choosing an attack in $V(\cal J)$ is always preferable as it gives a higher score to the attacker.

Let us denote $\preceq_{adv}$ the partial order relation on the set of vulnerability regions in which $\cal I \preceq_{adv} \cal J$ if and only if $V(\cal J) \neq \emptyset$ and $\cal I \subseteq \cal J$ \textit{i.e.} an attacker always prefers to attack in $V(\cal J)$ than in $V(\cal I)$. 

\paragraph{Adversarial lower semi-lattice.}

\begin{figure}[ht]
    \centering
    \captionsetup{width=1.0\linewidth}
      \includegraphics[scale=0.071]{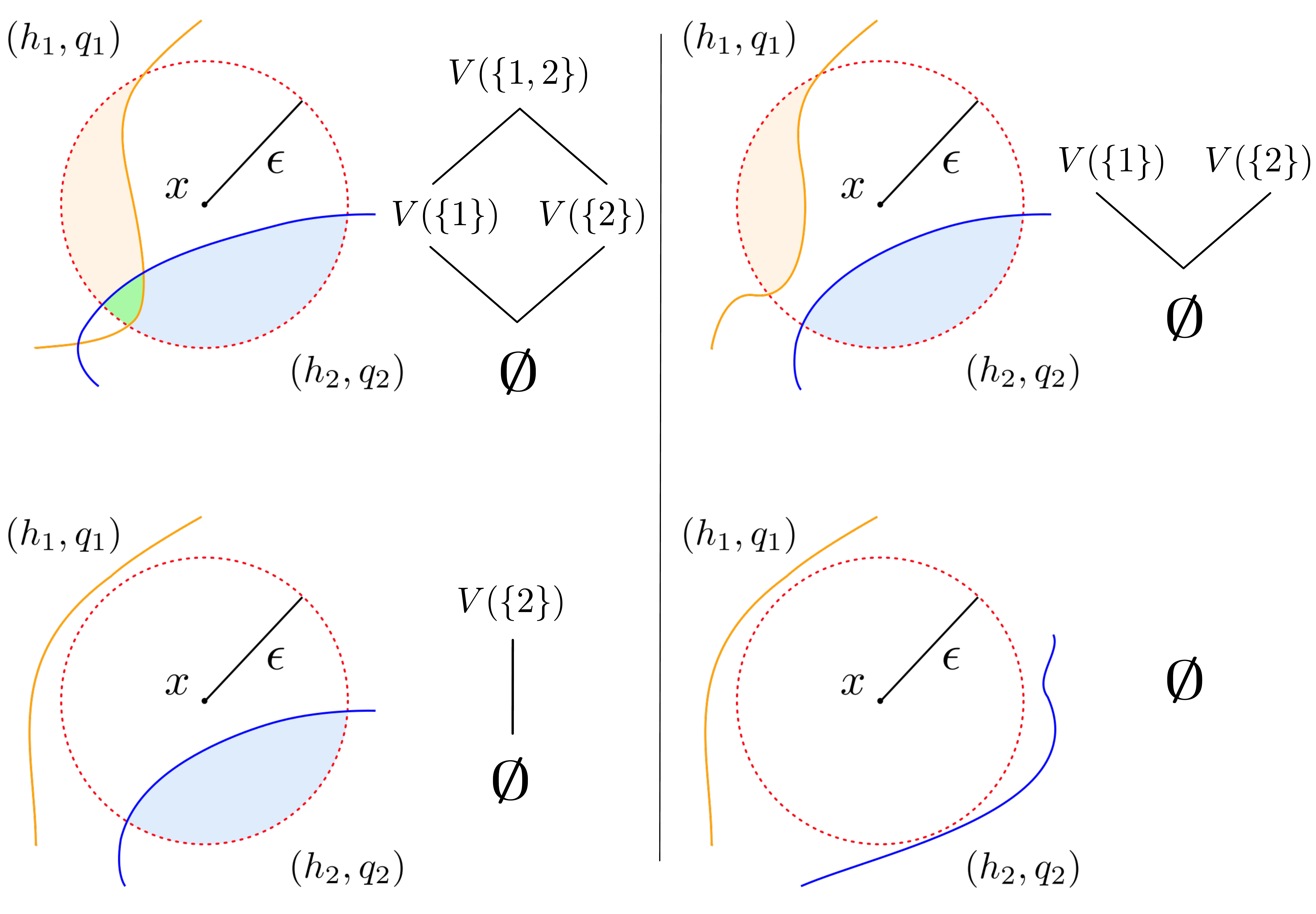}
    \caption{Examples of adversarial semi lattice around a point $x$ for the configurations described in Figure \ref{two_classifiers_regions}}
    \label{adv_semilattice}
\end{figure}

The order relation $\preceq_{adv}$ induces a \textit{lower semi-lattice structure} \cite{davey2002introduction} in the set of vulnerability regions or their corresponding index subsets (See Figure \ref{adv_semilattice}) where the empty set at the bottom corresponds to the non-vulnerability region in $B^\epsilon(x)$ and then, as we mount in the semi lattice, each vulnerability region gives a strictly better score to the attacker. More over, the \textit{maximal vulnerability regions} defined in Section \ref{sec:geometry} correspond to the $\preceq_{adv}$-maximal elements of the semi-lattice.

Having the semi-lattice object in mind, we can highlight a few important things:
\begin{itemize}
    \item The adversarial lattice can be any sub lower semi lattice of the power set $\mathcal{P}(\lbrace m \rbrace)$, depending on the arrangement of the classifiers $h_i$ inside the $\epsilon$-ball around $x$. An attack should be able to perform well in any of these configurations.
    \item The optimal attack for the mixture is inside one of the (possibly many) maximal regions of the semi lattice. Maximality is weaker than optimality when the number of maximal elements is greater than one.
    \item Effectiveness in the semi lattice translates to finding an adversarial attack in some element (at any level) of the lattice that is not the empty set at the bottom, whenever the lattice is not trivial. With the lattice in mind, it becomes even more evident that effectiveness is weaker than maximality and that the gap becomes potentially larger when the number of classifiers grows.
\end{itemize}

\paragraph{Climbing the semi-lattice.}

The main idea behind LCA if that to arrive at a maximal region, one can climb one level of the adversarial semi-lattice at a time. Figure \ref{ourattack_latticemount} shows the lattice mount behavior we want to achieve. In the rest of the section, we will explain our approach in two steps. First, in Section \ref{sec:attack_blc}, we develop an attack algorithm when dealing with binary linear classifiers where guarantees about maximality can be given. There, we will introduce the two components that make up our proposed attack: an {\em intersection finder} procedure and a {\em navigation mechanism}. Then, in Section \ref{sec:attack_multi}, we adapt the attack algorithm for multi-class differentiable classifiers like neural networks based on the ideas developed in the binary linear setting. 

\renewcommand\figscale{0.16}
\begin{figure}[ht]
    \centering

      \includegraphics[scale=\figscale]{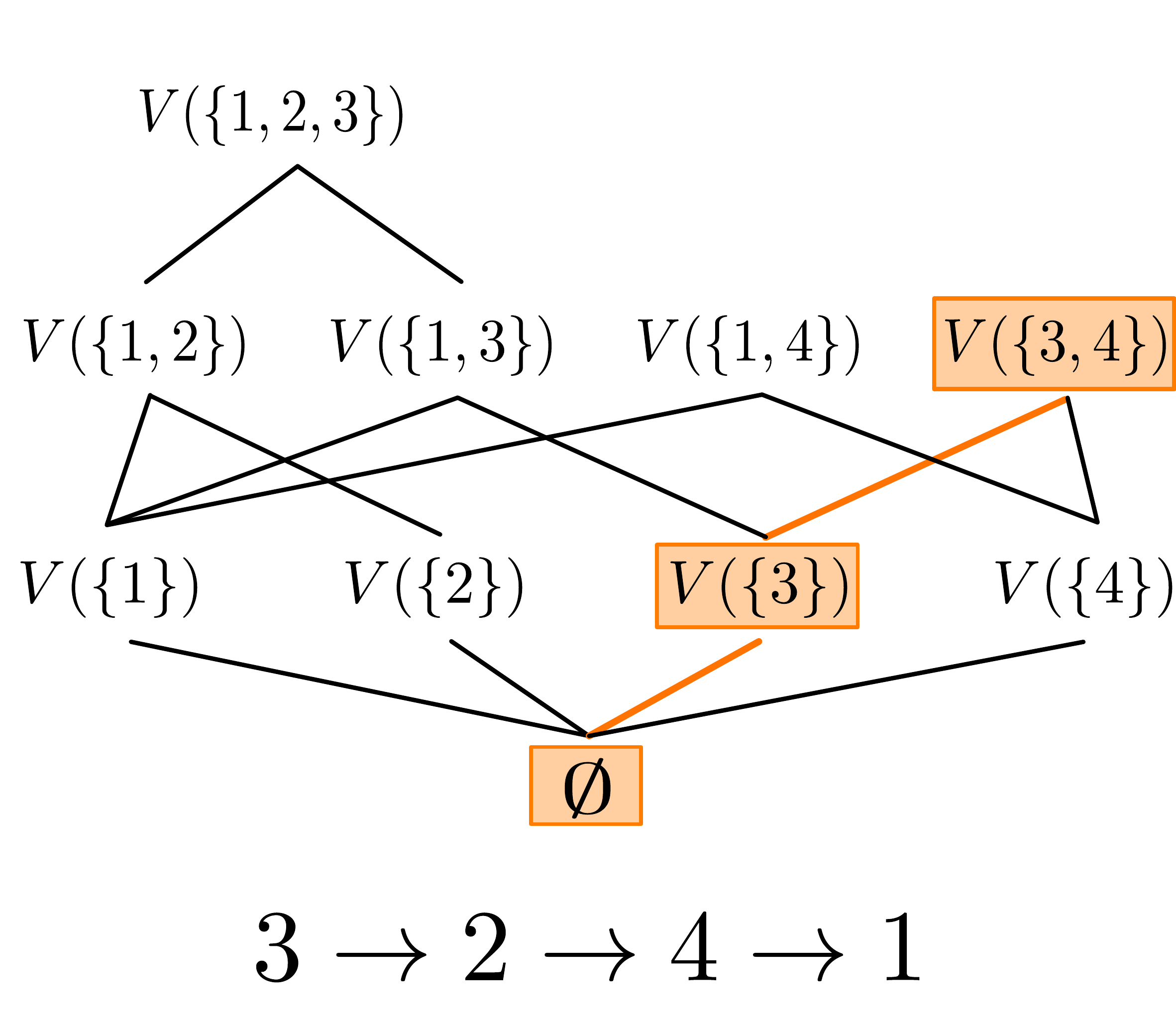}
    \caption{Main idea behind $LCA$. Given an ordering of the classifiers, we mount the lattice by adding one classifier to the pool of attacked ones at each step. In this example, we attack first the classifier $h_3$ successfully. Then we try to add $h_2$ without success ($h_3$ and $h_2$ have no common vulnerability region). Then we add $h_4$ with success and finally $h_1$ without success. In the end, the attack was able to attack $h_3$ and $h_4$ simultaneously getting a score of $q_3 + q_4$. The other maximal regions in this example are $V(\{1, 2, 3\})$ and $V(\{1, 4\})$.}
    \label{ourattack_latticemount}
\end{figure}

\subsection{Binary linear classifiers} \label{sec:attack_blc}

In this section, we consider class labels $y \in \lbrace -1, 1 \rbrace$ and classifiers $h : \mathbb R^d \rightarrow \lbrace -1, 1 \rbrace$ of the form $h(x) = \mathrm{sign}(f(x))$ for some linear function $f : \mathbb R^d \rightarrow \mathbb R$. In this setting, $h$ correctly classifies the data point $(x,y)$ if $f(x) \cdot y > 0$, \textit{i.e.} they have the same sign. Therefore, attacking $h$ translates to minimizing $f(x) \cdot y$. The optimal attack direction and margin to the decision boundary of a single linear classifier are known for both $\ell_2$ and $\ell_{\infty}$ norms \cite{dbouk2022arc}. 

\paragraph{Intersection finder.}
The first component of the algorithm is a procedure to attack a subset of classifiers $\cal I$ at the same time and find $x + \delta \in V(\cal I)$ if $V(\cal I) \neq \emptyset$. We will refer to this component as {\em intersection finder}, and is equivalent to the \textit{membership oracle} in \cite{boley2010listing} as we expect it to return a perturbation that allows us to check if \cal I belongs to the adversarial lattice or not. 

\subparagraph{Reverse hinge loss as intersection finder mechanism.}
As in \cite{perdomosinger}, we consider the {\em reverse hinge loss} $\ell_{rev}(y\cdot f(x)) = \max \left( y\cdot f(x), 0 \right)$. In the binary linear classifier setting, this function is convex and equal to $0$ if and only if $h$ misclassifies $x$.

Now, if we consider a set of classifiers $\bold h = \{ h_i\}_{i \in \cal I}$, the attacker would like to minimize $f_i(x) \cdot y$ for all $i \in \cal I$. In order to do so, one can minimize the sum of reverse hinge losses $SRH(\cal I, \bold h, x, y) =  \frac{1}{\lvert \cal I \rvert} \sum_{i \in \cal I} \ell_{rev}(y \cdot f_i(x))$ with projected gradient descent (PGD). Note that $SRH(\cal I, \bold h, x', y) =  0$ if and only if all the $h_i$, $i \in \cal I$ are fooled at the same time by $x'$. Also, if a classifier $h_i$ is already fooled, its loss term is not taken into account in the sum, only reappearing if at some step of the optimization procedure the new perturbation no longer fools it. This is in contrast with APGD \cite{pinot2020randomization}, which at all times attacks all the classifiers.

When all $f_i$ are linear functions of $x$, we get the condition we want for an {\em intersection finder}: $SRH$ is a convex function of $x$, and if all $h_i$ can be attacked at the same time, the sum has a global minimum $x'$ with value $0$. In this setting, running projected gradient descent with the correct parameters on the sum of reverse hinge will converge to some $x''$ that fools all the $h_i$ at the same time, see \cite[Theorem 3]{perdomosinger}.

\paragraph{Navigation mechanism.}
The second component of the algorithm corresponds to the capacity of an attacker to choose in the best way possible which classifiers to attack in order to efficiently navigate the adversarial lattice. Ideally, if we have a good {\em intersection finder}, we could test all possible subsets $\cal I \subseteq \left[ m \right]$ and find the optimal solution, but this becomes unfeasible for large values of $m$. Many algorithms could be adapted for the task of navigating the lattice, like \verb|Apriori|\cite{agrawal1994fast}, \verb|MaxMiner| \cite{bayardo1998efficiently} or \verb|AllMSS| \cite{gunopulos1997discovering}, but the time they require may be too large for our problem. For this reason, we use a much simpler navigation mechanism, akin to \verb|A_Random_MSS| in \cite{gunopulos1997discovering}.

Our navigation mechanism, linear in $m$, consists of keeping a pool of fooled classifiers and trying to expand the pool by adding one classifier at a time in a fixed order. The newly added classifier will be kept in the pool if it can be fooled with all ancient members of the pool at the same time. If this is not the case, we discard it and keep the old pool that is guaranteed to be attacked simultaneously. 

Note that the order in which we consider the classifiers is a parameter of the algorithm. Similar to \cite{dbouk2022arc}, we find that the heuristic of considering the classifiers in decreasing order of their associated probability $q_i$ yields a good performance. For example, in the case $m=2$, it ensures that LCA is optimal. The pseudocode is shown in Algorithm \ref{alg:binary_linear_attack}.

As we are in the binary linear classifier setting, we can set $T$ and $\eta$ so that PGD for the sum of reverse hinge losses is a perfect intersection finder \cite{perdomosinger} (more discussion on the Appendix). Under these assumptions, in line \ref{algo:blc:attack_current_pool} we will always find an attack $x + \hat{\delta} \in V(\cal I)$ if $V(\cal I) \neq \emptyset$. If the {\em intersection finder} is guaranteed to find such attack, we are also guaranteed to mount up the adversarial semi-lattice on some branch that is determined by the order in which we considered the classifiers. Figure \ref{ourattack_latticemount} shows the effect of the order in the outcome of the algorithm and how it will select a branch of the semi-lattice.

\begin{algorithm}[!t]
\caption{$LCA$ for binary linear classifiers}\label{alg:binary_linear_attack}
\begin{algorithmic}[1]
\REQUIRE Set $\bold h$ of $m$ binary linear classifiers in some order $\{h_1, \cdots, h_m\}$, starting point $(x, y)$. $T$ number of iterations and $\eta$ step size for PGD.
\ENSURE $\delta$ adversarial perturbation in some maximal region $V(\cal I)$ 
\STATE Initialize pool $\cal I = \emptyset$, $\delta = 0$
\FOR{$ k = 1, 2 \cdots, m$}\label{algo:blc:forloop}
 \STATE $\cal I = \cal I \cup \{ k \}$  \Comment{Add $h_k$ to the pool} \label{algo:blc:add_to_pool}
 \STATE Attack $SRH(\cal I, \bold h, x + \delta, y)$ with $PGD \left( T, \eta \right)$ to find new perturbation $\hat{\delta}$  \label{algo:blc:attack_current_pool}
 \IF{$SRH(\cal I, \bold h, x + \hat{\delta}, y) = 0$ \textit{i.e.} succeeded}
    \STATE $\delta = \hat{\delta}$ \Comment{Update current attack, keep $k$ in pool} \label{algo:blc:update_point}
 \ELSE{} 
    \STATE $\cal I = \cal I \setminus \{ k \}$ \label{algo:blc:reset_pool}
\ENDIF
\ENDFOR
\STATE \textbf{return} $x + \delta$\label{algo:blc:return-end}
\end{algorithmic}
\label{algo_blc}
\end{algorithm}

We end this part by stating the maximality of LCA in the binary linear setting (proof in Appendix \ref{app:proofs}).

\begin{theorem}[LCA is maximal in the binary linear setting] \label{thm:binary_maximality}
Let ${\bf m_h^q}$ be a mixture of binary linear classifiers. Fix $\epsilon > 0$ the attack budget. Then for any $(x, y)~\in~\mathbb{R}^d~\times~\{-1,1\}$, there exist parameters $T$ and $\eta$ for the inner PGD such that Algorithm \ref{alg:binary_linear_attack} returns an adversarial example $x'$ that is in a maximal vulnerability region of $\bold h$.
\end{theorem}

\subsection{Multi-class differentiable classifiers} \label{sec:attack_multi}

The ideas developed in Section \ref{sec:attack_blc} for the binary linear classifiers need to be adapted to the general multi class case with differentiable classifiers, like neural networks, because we can no longer have guarantees about the optimality of intersection finders like PGD on the sum of reverse hinge losses. Moreover, the reverse hinge loss was defined for binary classifiers and not for multi-class classifiers, so it needs an adaptation.

Given a non-linear differentiable classifier $h: \mathbb R^d \to \Delta^k$ and a point $(x, y)$, we are going to choose a target class $y_{adv} \in \left[ k \right]$ and turn the situation into a binary classification problem trying to push $h$ to predict $y_{adv}$. To do so, we minimize the reverse hinge loss of the margin between the probits (or logits) of the correct and the target class, \textit{i.e.} $\ell_{rev}(h(x)(y) - h(x)(y_{adv}), x, y)$ with $h(x)(j)$ denoting the $j$-th component of the vector $h(x)$. Minimizing this margin will, intuitively, push $\ell_{rev}(h(x)(y) )$ down and $\ell_{rev}(h(x)(y_{adv}))$ up. 

There are different ways to choose $y_{adv}$. One simple way is to choose the class $y_{adv} \neq y$ with the largest logit \cite{perdomosinger}. Another way is to approximate the decision boundaries for each class $j \neq y$ with linear functions and compute the distance to the linearized boundaries explicitly to choose the closest one. This is the approach followed in \cite{dbouk2022arc} to develop ARC.

Our proposed attack is presented in Algorithm \ref{alg:multiclass_attack}. The main difference w.r.t Algorithm \ref{alg:binary_linear_attack} is that there is no guarantee on the convergence of the attack to $SRH(\cal I, \bold h, x + \delta, y)$ in line \ref{algo:attack_current}. As we can no longer guarantee the optimality of the intersection finder, we change the criteria to update the pool of classifiers. In the binary linear classifier case, we enforced a strong maximality constrain, while here we impose a more flexible criterion based on the obtained score (loss). Each time we find a perturbation $\delta$ with a higher loss for the total mixture, we keep it and update the pool to be the classifiers fooled by $x +\delta$.

\begin{algorithm}[!t]
\caption{$LCA$ for multi-class classifiers}\label{alg:multiclass_attack}
\begin{algorithmic}[1]
\REQUIRE Set of $m$ classifiers $\bold h$ in some order $\{h_1 , \cdots, h_m \}$ and their probabilities $\bold q$. Starting point $(x, y)$. $T$ number of iterations and $\eta$ step size for PGD.
\STATE Initialize pool $\cal I = \emptyset$, $\delta = 0$
\FOR{$ k = 1, 2 \cdots, m$}\label{algo:forloop}
 \STATE $\cal I = \cal I \cup \{ k \}$ \label{algo:add_to_pool}
    
    \STATE Attack $SRH(\cal I, \bold h, x + \delta, y)$  with $PGD(T, \eta)$ to produce new perturbation $\hat{\delta}$ \label{algo:attack_current} 
    \IF{$\hat{\delta}$ produces a better score than $\delta$}
        \STATE $\delta = \hat{\delta}$ \label{algo:update_point}
    \ENDIF
    \STATE Recompute pool $\cal I$ according to current $\delta$       
\ENDFOR
\STATE \textbf{return} $x + \delta$\label{algo:return-end}
\end{algorithmic}
\label{algo_multiclass}
\end{algorithm}

\section{Experiments}

\subsection{Experiments on synthetic data} \label{toy_examples}
\begin{figure}[ht]
\centering
    \minipage{0.45\columnwidth}
    \centering
        \includegraphics[scale=0.25]{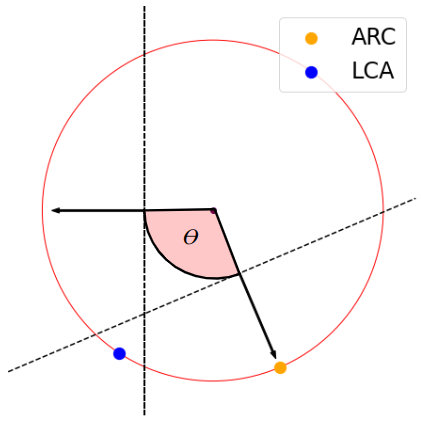}
    \endminipage\hfill
    \minipage{0.45\columnwidth}%
    \centering
        \includegraphics[scale=0.25]{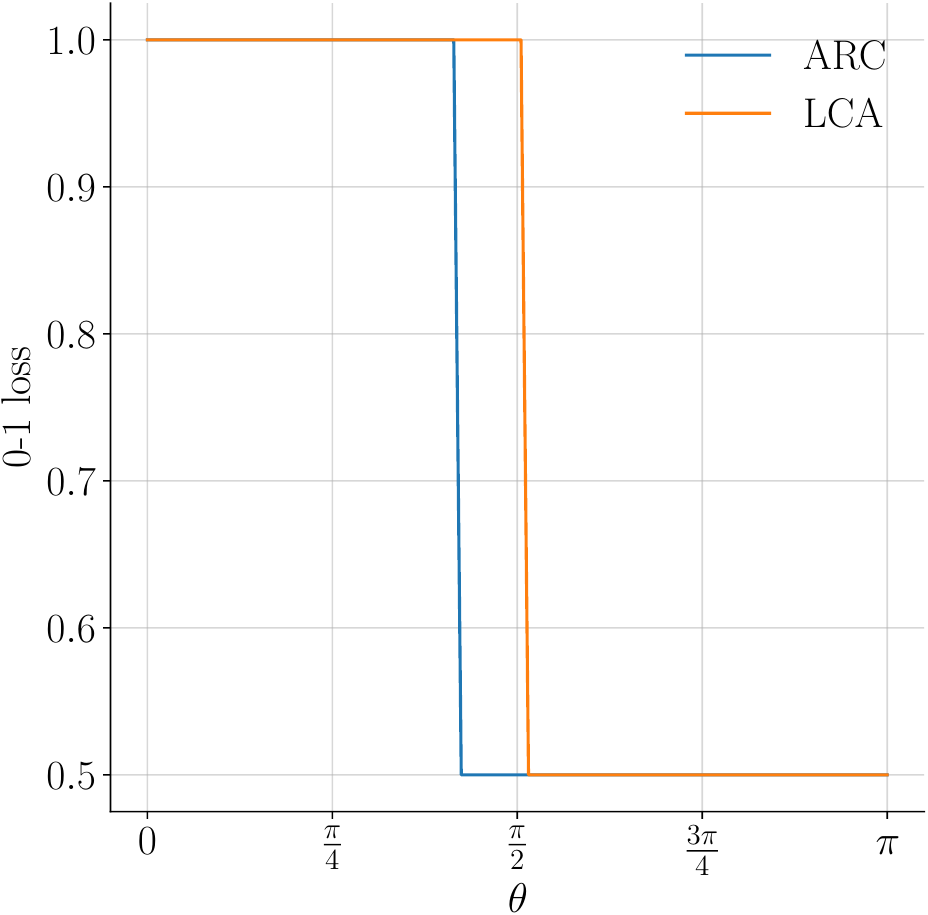}
    \endminipage
    \centering
    \captionsetup{width=1.0\linewidth}
    \caption{Left:  One example of the situation with a fixed $\theta$ where ARC fails to find the intersection. Right: Score obtained by LCA and ARC w.r.t the angle between the two normal vectors describing the linear classifiers. Note: a score of 1 means the two classifiers of the mixtures were successfully attacked (with the same attack), whereas a score of 0.5 means only one classifier out of two was successfully attacked. }
    \label{example_angle}
\end{figure}

We first focus on the simple task of binary classification using mixtures of two linear classifiers. In this setting, non-maximal attacks such as ARC can fail to discover an adversarial example when the common vulnerability region is too small. Instead, maximal attacks such as LCA with the right parameters are guaranteed to find it. To illustrate this difference, we vary the size of the common vulnerability region, by varying the angle $\theta$ between the normal vectors defining the two linear classifiers. Increasing the angle $\theta$ from 0 to $\pi$ will reduce the size of the common vulnerability region, as illustrated in Figure~\ref{example_angle} (left).
We then measure the score of the attack for various angles $\theta$ and report it in Figure~\ref{example_angle} (right). 

As we can see, the score of ARC drops to 0.5 before the score of LCA. Furthermore, we can verify that LCA is optimal in the sense that the score of the attacks drop only where there are no more common vulnerability region.

\renewcommand\figscale{0.15}

\begin{figure}[ht]

    \centering
      \includegraphics[scale=0.5]{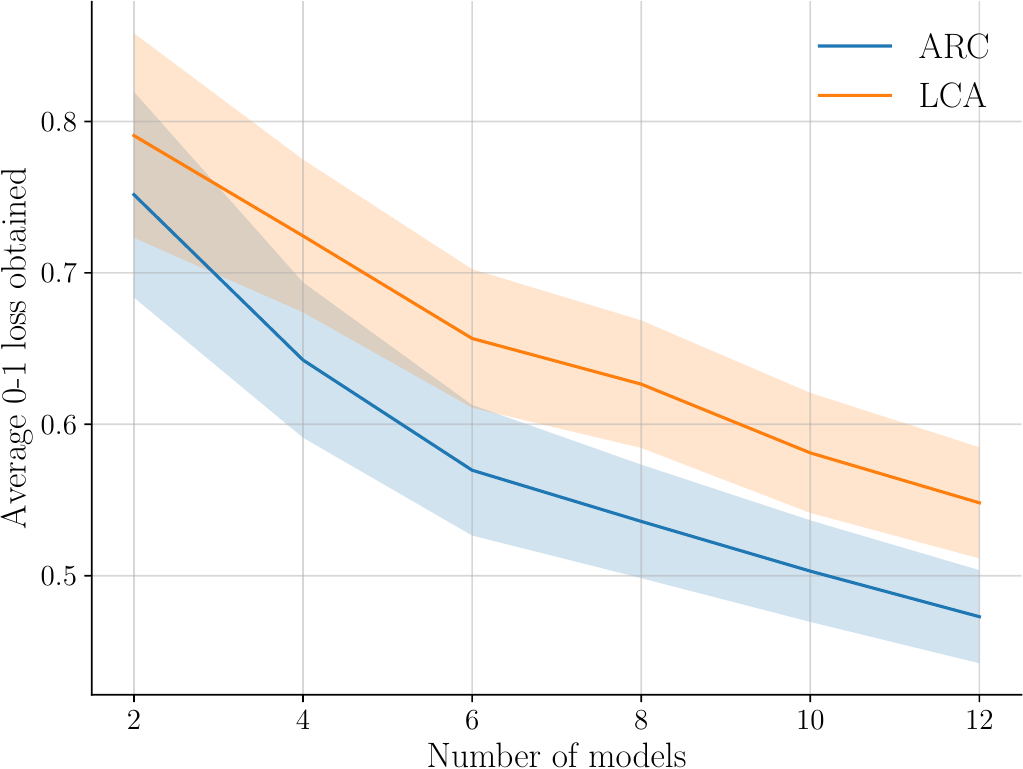}
    \caption{Mean score obtained by ARC and LCA w.r.t the number of models in $\mathbb R^{256}$. The score obtained by the attacker is the $0\mbox{}~-1$ loss (larger is better). For the sampling of linear classifiers, we sampled normal vectors uniformly in the sphere, and biases from $\cal N(0.5, 0.5^2)$. The colored regions around lines correspond to 0.25 standard deviations, where 0.25 was selected for visual purposes only.}
    \label{dim256_ARC_LC}
\end{figure}

In the next experiment, we evaluate the efficiency of the attacks against large mixtures. To do so, we measure the efficiency of each attack against mixtures with an increasing number of models. We consider $m$ linear classifiers sampled around a fixed point $x$ in dimension $d = 256$. Each linear model is represented by its normal vector in $\mathbb R^{256}$ sampled uniformly on the unit sphere and its bias in $\mathbb R$ sampled from a normal distribution $\cal N (0.5, 0.5^2)$. We then measure the adversarial loss of the mixture against each attack. We repeat this experiment $500$ times for each value of $m$ and report the average adversarial loss for each attack in Figure~\ref{dim256_ARC_LC}.

In this first experiment, we can see that on average, LCA is a more damaging attack than ARC is, and more importantly, that the performance gap between the two attacks increases with the number of models in the mixtures. We also found that, although LCA is always better on average, the performance of both  attacks is sensitive to the distribution used to sample classifiers. Finally, we also found that the parameter $T$ of the inner PGD (see~Algorithm~\ref{alg:binary_linear_attack}) has a strong impact on the execution times. See Appendix \ref{app:toy}.

\subsection{CIFAR-10 and CIFAR-100}

In this section, we test the attacks against mixtures built using DVERGE \cite{yang2020dverge} and GAL \cite{kariyappa2019gal}, two different ensemble training methods that were developed to induce diversity in the models and improve robustness. We used the implementation given in \cite{yang2020dverge} to train the classifiers on the CIFAR-10 and CIFAR-100 datasets \cite{krizhevsky2009cifar}. In all settings, we denote by Baseline the naive mixture built from $m$ classifiers trained independently on clean images.

We tested mixtures with an increasing number of classifiers to see the effect this parameter has on the attack performance. For comparison, we also attacked the models with APGD \cite{pinot2020randomization}, and ARC \cite{dbouk2022arc}. All three attacks can be easily adapted to both $\ell_2$ and $\ell_{\infty}$ norms, so we perform attacks on both norms with the standard values $\epsilon = 0.5$ for the $\ell_2$ threat model, and $\epsilon = \frac{8}{255}$ for the $\ell_{\infty}$ threat model. All the parameters for the training as well as the attacks are detailed in Appendix \ref{app:cifar}. Table~\ref{table:cifar10_linf} presents the empirical adversarial risk (over 10~000 samples) for CIFAR-10 and the $\ell_{\infty}$ threat model.

 First, we can see that the baselines are all very vulnerable:  both APGD and LCA are able to bring down to 0.0, and ARC is also able to bring the accuracy down to almost 0. For the DVERGE models the robustness is good when the classifier is tested against APGD, and even against ARC, but its accuracy is brought down to almost 0 when it is tested against LCA. This confirms that APGD (and ARC to a more limited extent) gives a false sense of robustness \cite{dbouk2022arc} discussed earlier in this paper. For the GAL models that are less performant, ARC seems to have a harder time finding perturbations that are adversarial to all the models. APGD performs better and again and LCA is performing far better than the other attacks in this setting.

\begin{table}
\caption{Robustness results in CIFAR-10, $\ell_{\infty}$ threat model. Accuracy is expressed as a percentage. Details in Appendix \ref{app:cifar}} \label{table:cifar10_linf}
\begin{center}
\begin{tabular}{l|llll}
\hline
\textbf{Model}  & \textbf{Natural} & \textbf{APGD} & \textbf{ARC} & \textbf{LCA} \\
\hline
Baseline(3) & 91.8 & 0.0 & 0.9 & 0.0 \\ 
Baseline(5) & 91.9 & 0.0 & 1.5 & 0.0 \\ 
Baseline(8) & 91.8 & 0.0 & 1.7 & 0.0 \\ 
Baseline(12) & 90.4 & 0.0 & 2.0 & 0.0 \\ 
\hline
Dverge(3) & 90.1 & 33.3 & 14.4 & 0.1 \\ 
Dverge(5) & 89.8 & 31.3 & 26.3 & 0.6 \\ 
Dverge(8) & 88.7 & 25.2 & 33.6 & 2.6 \\ 
\hline
GAL(3) & 85.1 & 5.0 & 4.2 & 0.0 \\ 
GAL(5) & 84.7 & 6.1 & 13.7 & 0.0 \\ 
GAL(8) & 83.9 & 4.2 & 19.0 & 0.1 \\ 
\hline
\end{tabular}
\end{center}
\end{table}

Table \ref{table:cifar10_l2} changes the threat model to $\ell_2$. Again, baselines are 100\% vulnerable, and ARC and LCA can bring the accuracy near 0. We can see that the number of models considerably affects the performance of attacks, specially that of ARC and LCA. As in the last setting, LCA also performs better in this setting. 

\begin{table}
\caption{Robustness results in CIFAR-10, $\ell_{2}$ threat model. Accuracy is expressed as a percentage. Details in Appendix \ref{app:cifar}}
\label{table:cifar10_l2}
\begin{center}
\begin{tabular}{l|llll}
\hline
\textbf{Model}  & \textbf{Natural} & \textbf{APGD} & \textbf{ARC} & \textbf{LCA} \\
\hline
Baseline(3) & 91.8 & 4.8 & 0.5 & 0.0 \\ 
Baseline(5) & 91.9 & 3.1 & 0.5 & 0.0 \\ 
Baseline(8) & 91.8 & 2.4 & 0.5 & 0.0 \\ 
Baseline(12) & 90.4 & 3.3 & 0.9 & 0.1 \\ 
\hline
Dverge(3) & 90.1 & 47.4 & 16.2 & 9.0 \\ 
Dverge(5) & 89.8 & 51.4 & 26.9 & 17.2 \\ 
Dverge(8) & 88.7 & 51.4 & 38.2 & 29.0 \\ 
\hline
GAL(3) & 85.1 & 24.2 & 6.4 & 3.0 \\ 
GAL(5) & 84.7 & 28.1 & 9.8 & 4.4 \\ 
GAL(8) & 83.9 & 26.9 & 13.1 & 6.4 \\ 
\hline
\end{tabular}
\end{center}
\end{table}

\begin{table}
\caption{Robustness results in CIFAR-100, $\ell_{\infty}$ threat model. Accuracy is expressed as a percentage. Details in Appendix \ref{app:cifar}} \label{table:cifar100_linf}
\begin{center}
\begin{tabular}{l|llll}
\hline
\textbf{Model}  & \textbf{Natural} & \textbf{APGD} & \textbf{ARC} & \textbf{LCA} \\
\hline
Baseline(3) & 64.6 & 0.3 & 1.2 & 0.0 \\ 
Baseline(5) & 64.6 & 0.2 & 2.5 & 0.0 \\ 
Baseline(8) & 64.8 & 0.1 & 3.2 & 0.0 \\ 
Baseline(12) & 64.6 & 0.1 & 3.7 & 0.0 \\ 
\hline
Dverge(3) & 60.7 & 8.9 & 8.5 & 0.1 \\ 
Dverge(5) & 59.7 & 12.0 & 19.2 & 0.5 \\ 
Dverge(8) & 61.1 & 13.1 & 24.7 & 1.6 \\ 
\hline
GAL(3) & 45.6 & 0.7 & 2.7 & 0.0 \\ 
GAL(5) & 40.5 & 0.5 & 5.3 & 0.0 \\ 
GAL(8) & 47.6 & 0.7 & 9.4 & 0.0 \\ 
\hline
\end{tabular}
\end{center}
\end{table}

\begin{table}
\caption{Robustness results in CIFAR-100, $\ell_{2}$ threat model. Accuracy is expressed as a percentage. Details in Appendix \ref{app:cifar}}
\label{table:cifar100_l2}
\begin{center}
\begin{tabular}{l|llll}
\hline
\textbf{Model}  & \textbf{Natural} & \textbf{APGD} & \textbf{ARC} & \textbf{LCA} \\
\hline
Baseline(3) & 64.6 & 3.4 & 1.2 & 0.0 \\ 
Baseline(5) & 64.6 & 3.1 & 1.5 & 0.0 \\ 
Baseline(8) & 64.8 & 2.8 & 2.0 & 0.1 \\ 
Baseline(12) & 64.6 & 2.3 & 2.1 & 0.2 \\ 
\hline
Dverge(3) & 60.7 & 17.6 & 3.6 & 1.7 \\ 
Dverge(5) & 59.7 & 22.1 & 8.1 & 4.7 \\ 
Dverge(8) & 61.1 & 27.4 & 15.2 & 11.0 \\ 
\hline
GAL(3) & 45.6 & 4.3 & 1.7 & 0.3 \\ 
GAL(5) & 40.5 & 4.2 & 2.1 & 0.4 \\ 
GAL(8) & 47.6 & 6.5 & 3.3 & 0.8 \\ 
\hline

\end{tabular}
\end{center}
\end{table}

\paragraph{Final discussion}

In conclusion, LCA performs consistently better against mixtures of classifiers across multiple datasets and threat models. It is worth distinguishing two particular regimes: the first regime is when all models can be attacked simultaneously (i.e. when the common vulnerability region is not empty). In this case, APGD performs well, and LCA is able to match this performance while ARC typically degrades, specially with large mixtures. The second, more interesting regime, is when there is no common vulnerability region. In this case APGD performs poorly, and ARC, which was built to address this flaw, works better. However, unlike LCA, it does not guarantee maximality and LCA (who does) performs better in practice. 

\section{Related work}

\paragraph{Attacking a mixture of classifiers.}
Mixtures were introduced in \cite{pinot2020randomization} as a way to decrease the worst-case theoretical robustness guarantee of a single classifier that faces a regularized adversary. This idea lead to BAT, an algorithm that constructs a mixture via a boosting like process to improve robustness. This method and also the one proposed in \cite{meunier2021mixed} were tested against attacks used for individual classifiers (PGD\cite{madry2017towards}  or C\&W\cite{carlini2017towards}) that were \textit{adapted} to the mixture setting by considering the deterministic expected model. This was shown to be problematic \cite{dbouk2022arc} because the attack would fail to find a perturbation, even when it was rather easy to attack at least one of the models, leading to an overly-optimistic robustness estimate. This property, called \textit{inconsistency} in \cite{dbouk2022arc}, was the core idea behind their proposed attack ARC (Attacking Randomized ensembles of Classifiers).

The work of \cite{perdomosinger} also studies the problem of attacking multiple classifiers simultaneously. Under the lens of that work, we also aim at proposing a \textit{best response oracle} that can produce an optimal perturbation for a fixed set of classifiers given their weights. Their work explains the setting from a game theory perspective and proposes a way to attack multiple classifiers.

\subsection{Relation with the problem of finding frequent item sets.}
The problem of finding maximal elements in a lattice has been studied in the domain of \textit{data mining}. Algorithms like \verb|Apriori| \cite{agrawal1994fast} or \verb|MaxMiner| \cite{bayardo1998efficiently} were proposed to find \textit{frequent item sets} in transaction databases, which can be stated as finding maximal elements in a lattice. In \cite{boley2010listing}, the authors create a very general framework, from which we can also instantiate our problem and that of \textit{frequent item set mining}. 

In \cite{gunopulos1997discovering}, the authors propose algorithm \verb|All_MSS| to enumerate \textit{all} maximal elements of what they call a \textit{theory}. Our proposed LCA attack can be seen as a parallel of algorithm \verb|A_Random_MSS| \cite{gunopulos1997discovering}, which performs exactly one random lattice climb, finding one maximal item. It is important to highlight that for our particular problem applied to general models like neural network, the attack should be very fast, specially if it is used for some training method that involves some sort of adversarial training. This excludes the methods that enumerate all maximal regions, and leaves us with simpler methods that try to at least arrive at one maximal region.

\paragraph{Ensembles and robustness.}
The use of ensembles has also been proposed as a way to improve robustness to adversarial attacks. Methods like ADP \cite{pang2019adp}, GAL \cite{kariyappa2019gal}, EMPIR \cite{empir}, GPMR \cite{dabouei2020gpmr}, LIT \cite{ross2020lit},
TRS \cite{yang2021trs}, DVERGE \cite{yang2020dverge} or MRBOOST \cite{zhang2022building} all try to design a training procedure able to create ensembles that are, as a whole, more robust to adversarial attacks.
It is worth noting that ensembles and mixtures are different extensions of a family of classifiers, and attacking these two types of models poses two different optimization problems. In this work, we have focused on mixtures, and we have built them using ensemble training procedures following the hypothesis that diverse models that are not simultaneously vulnerable should also yield robust mixtures. The problem of training robust mixtures is still an open problem to the best of our knowledge.

\section{Conclusion}
We analyzed the problem of attacking a mixture of classifiers. This setting is different from the deterministic one and requires attacks adapted to the combinatorial nature of the problem. This complexity that arises as the number of models increases makes attacking a mixture more challenging, and therefore using mixtures could be a way to achieve better robustness to adversarial attacks. 
Inspired by recent approaches \cite{perdomosinger, dbouk2022arc} we presented a framework that simplifies the understanding of the problem of attacking a mixture of classifiers. This framework also allows us to analyze existing attacks and their properties. Building on previous work, we proposed an attack (LCA) that has stronger guarantees in the binary linear classifier case than existing ones. This superiority was confirmed by comparing LCA with ARC on synthetic data that was particularly crafted to maintain the characteristic properties of each one. In the case of general models like neural networks, the intuitions developed for the binary linear classifiers setting allowed us to propose an attack that is better than existing ones (APGD and ARC).


\bibliography{ecai}

\newpage

\onecolumn

\appendix

\section{Proofs} \label{app:proofs}

\textbf{Theorem (Hardness of attacking linear classifiers)}. Consider a binary classification
setting. Given a labeled point $(x,y)$, a set of $m$ linear classifiers
$x\mapsto\mathds{1}\left\{ \theta_{i}^{\top}x+b_{i}\ge0\right\} $
where $\theta_{i},b_{i}\in\mathbb{R}^{d+1}$, a uniform mixture ${\bf m}$
composed of these linear classifiers, a noise budget $\epsilon>0$
and a value $\beta>0$, the problem of checking if there exists $\delta\in B^{\epsilon}(x)$
such $\ell^{0-1}\left({\bf m},x+\delta,y\right)\ge\beta$
is NP-hard.

\noindent
\textit{Proof:}

To show NP-hardness of our problem, we will use a reduction involving
the NP-hard \emph{MaxFLS} problem defined as follows:
\begin{itemize}
\item \textbf{Input of MaxFLS: }A set of $m$ pairs $\left\{ \left(\theta_{i},b_{i}\right)\right\} _{i=1}^{m}$
where $\theta_{i}\in\mathbb{R}^{d}$and $b\in\mathbb{R}$, and a value
$\alpha>0$
\item \textbf{Output of MaxFLS:} \verb|Yes| if there exists a vector $\delta\in\mathbb{R}^{d}$
such that at least a fraction $\alpha$ of inequalities among the set
$\left\{ \delta^{\top}\theta_{i}+b_{i}\ge0:i\in[m]\right\} $ are
satisfied
\end{itemize}
Let us consider an instance $\left(\left\{ \left(\theta_{i},b_{i}\right)\right\} _{i=1}^{m},\alpha\right)$
of MaxFLS. The set $\left\{ \left(\theta_{i},b_{i}\right)\right\} _{i=1}^{m}$
defines an arrangement of hyper-planes in $\mathbb{R}^{d}$, and this
arrangement partitions the space $\mathbb{R}^{d}$ in regions. Let
$x=0$. Clearly, we can always find a value $\epsilon$ such that
each region has a non-empty intersection with the ball $B^{\epsilon}(x)$.
Let $y=0$ and ${\bf q}=\left(\frac{1}{m}\ldots\frac{1}{m}\right)$.
Define ${\bf m}$ as the mixture of binary classifiers $\left\{ x\mapsto\mathds{1}\left\{ \theta_{i}^{\top}x+b_{i}\ge0\right\} \right\} _{i\in[m]}$
with weights ${\bf q}$. Then, $\ell^{0-1}\left({\bf m},x+\delta,y\right)=\frac{1}{m}\sum_{i\in[m]}\mathds{1}\left\{ \delta^{\top}\theta_{i}+b_{i}\ge0\right\} $.
Clearly, the value $\ell^{0-1}\left({\bf m},x+\delta,y\right)$
is equal to the fraction of satisfied inequalities among the set $\left\{ \delta^{\top}\theta_{i}+b_{i}\ge0:i\in[m]\right\} $.
Thus, $\ell^{0-1}\left({\bf m},x+\delta,y\right)\ge\alpha$
if and only if at least $\alpha$ inequalities in MaxFLS are satisfied. Thus, if a polynomial time algorithm was able to solve this attacking problem, we could also use it to solve the MaxFLS problem in polynomial time, which is a contradiction, as MaxFLS is NP-hard. Thus, our problem is also NP-hard.
\begin{flushright}
$\square$
\end{flushright}

\begin{lemma}\label{lemma:climb_step}
Let ${\bf m_h^q}$ be a mixture of $m$ binary linear classifiers. Fix $\epsilon > 0$ the attack budget and a point $(x, y)$. If there exists an adversarial example $x' \in B^{\epsilon}(x)$ for all classifiers $h_i$ simultaneously, then there exist parameters $T$ and $\eta$ such that minimizing $SRH(\bold h, x, y)$ with PGD($T$, $\eta$) will produce some adversarial example $x''$ for all $h_i$ simultaneously.
\end{lemma}

\noindent
\textit{Proof (See \cite[Appendix F]{perdomosinger} and \cite{bubeck2015convex}):}

Note that the hypothesis that there exists $x'$ adversarial to all $h_i$ means that $SRH(\bold h, x', y) = 0$. This is a global minimum, as $SRH$ is a non-negative function.
Also note that $SRH$ as a function of $x$ can only take a finite number of values, the smallest positive one being $\frac{1}{m}$. By \cite[Theorem 3]{perdomosinger}, \cite[Theorem 3.2]{bubeck2015convex} with $\delta < \frac{1}{m}$, running PGD for $T > \epsilon^2 \cdot m^2$ steps with step size $\eta = \frac{\epsilon}{\sqrt{T}}$ returns a solution $x''$ such that $SRH(\bold h, x'', y) - SRH(\bold h, x', y) = SRH(\bold h, x'', y) < \frac{1}{m}$. This implies that $SRH(\bold h, x'', y) = 0$.
\begin{flushright}
$\square$
\end{flushright}

\noindent
\textbf{Theorem \ref{thm:binary_maximality} (LCA is maximal in the binary linear setting)}
Let ${\bf m_h^q}$ be a mixture of binary linear classifiers. Fix $\epsilon > 0$ the attack budget. Then for any $(x, y)~\in~\mathbb{R}^d~\times~\{-1,1\}$, there exist parameters $T$ and $\eta$ for the inner PGD such that Algorithm \ref{alg:binary_linear_attack} returns an adversarial example $x'$ that is in a maximal vulnerability region of $\bold h$.

\noindent
\textit{Proof:}

Suppose by contradiction that Algorithm \ref{alg:binary_linear_attack} returns $x'$ that is in some vulnerability region $V(\cal I)$ that is not maximal. 

By definition of maximality, this means there exists some $j \in [m] \setminus \cal I$ such that $V(\cal I \cup \{j\}) \neq \emptyset$.

Denote $\cal I_i$ the pool of fooled classifiers at each step $i$ of the $m$ steps in the outer loop of Algorithm \ref{alg:binary_linear_attack}, with $\cal I_0 = \emptyset$ and $\cal I_m = \cal I$. We have that by construction, $\cal I_{i} \subseteq \cal I_{i+1}$. At step $j-1$ of the algorithm, the pool $\cal I_{j-1}$ consisted of classifiers that were all vulnerable at the same time. Given that $\cal I_{j-1} \cup \{j\} \subset \cal I \cup \{j\}$, we have that $V(\cal I_{j-1} \cup \{j\}) \supset V(\cal I \cup \{j\}) \neq \emptyset$.

As $V(\cal I_{j-1} \cup \{j\}) \neq \emptyset$, at step $j$, Lemma \ref{lemma:climb_step} says that the PGD step for attacking the classifiers $\cal I_{j-1} \cup \{j\}$ simultaneously returns $x''$ that fools all of them, meaning that $j$ would be added to the pool, \textit{i.e} $\cal I_j = \cal I_{j-1} \cup \{j\}$. This implies that $j \in \cal I$, which is a contradiction.

This contradiction arises from supposing that $V(\cal I)$ was not maximal. Therefore, we can conclude that Algorithm \ref{alg:binary_linear_attack} returns $x'$ in a maximal vulnerability region.

\begin{flushright}
$\square$
\end{flushright}

\rule{\textwidth}{0.7pt}

\section{Toy examples details} \label{app:toy}
\subsection{ARC and LCA against two classifiers at the same distance from $x$ in an angle of $\theta$.}
For this experiment, the parameters of LCA are
\begin{itemize}
    \item Norm (threat model): $\ell_2$.
    \item Attack budget $\epsilon$: 1.
    \item Number of steps $T$ for inner PDG: 100.
    \item Step size $\eta$ for inner PDG: $\frac{\epsilon}{20}$.
\end{itemize}

\subsection{ARC and LCA against randomly sampled mixtures of different sizes}
For the plots presented in Section \ref{toy_examples} comparing ARC and LCA against mixtures with different number of classifiers, and in this appendix, the parameters for the LCA attack were:
\begin{itemize}
    \item Norm (threat model): $\ell_2$.
    \item Attack budget $\epsilon$: 1.
    \item Number of steps $T$ for inner PDG: 200.
    \item Step size $\eta$ for inner PDG: $\frac{\epsilon}{T}$.
\end{itemize}

Other important parameters and details of the simulation of linear classifiers were:
\begin{itemize}
    \item \verb|numpy| seed was set to 42
    \item {Weights $\bold q$ of each random mixture were also sampled randomly as follows:
        \begin{enumerate}
            \item Sample a vector $z$ of $m$ i.i.d standard normal, $z_i \sim \cal N(0,1)$.
            \item Compute the softmax with temperature $t=10$, \textit{i.e} $q_i = \frac{\exp{(z_i/10)}}{\sum_j \exp{(z_j/10)}}$.
        \end{enumerate}
    }
    \item To sample a mixture of $m$ models, we sampled $m$ unit vectors in $\mathbb{R}^{256}$, and $m$ biases from a normal distribution $\cal N(\mu, \sigma^2)$. In the rest of the appendix we present the same plot shown in Section \ref{toy_examples} with different values of $\mu$ and $\sigma$, which drastically changed the performance of both LCA and ARC. In all cases, LCA clearly better in average, except for a very particular case in which both classifiers perform equally good.
\end{itemize}

\subsection{Results of LCA and ARC with different number of classifiers with respect to the distribution of the bias of the linear classifiers}

Here we present the same plot as in Figure \ref{dim256_ARC_LC} but with a different distribution for the classifiers composing the mixtures. We changed the mean of the bias, which translates to the distance on average of each linear classifier to the center point $x$. We also changed the standard deviation of the bias. If it is very small, this means that with high probability all linear classifiers will be close to the center point. On the contrary, a big standard deviation means that the distance from each classifier to the center point can vary more. These two parameters affect the number and size of intersections of the linear classifiers inside the $\epsilon$-ball around $x$.

\begin{figure}[h]
    \centering
      \includegraphics[scale=0.5]{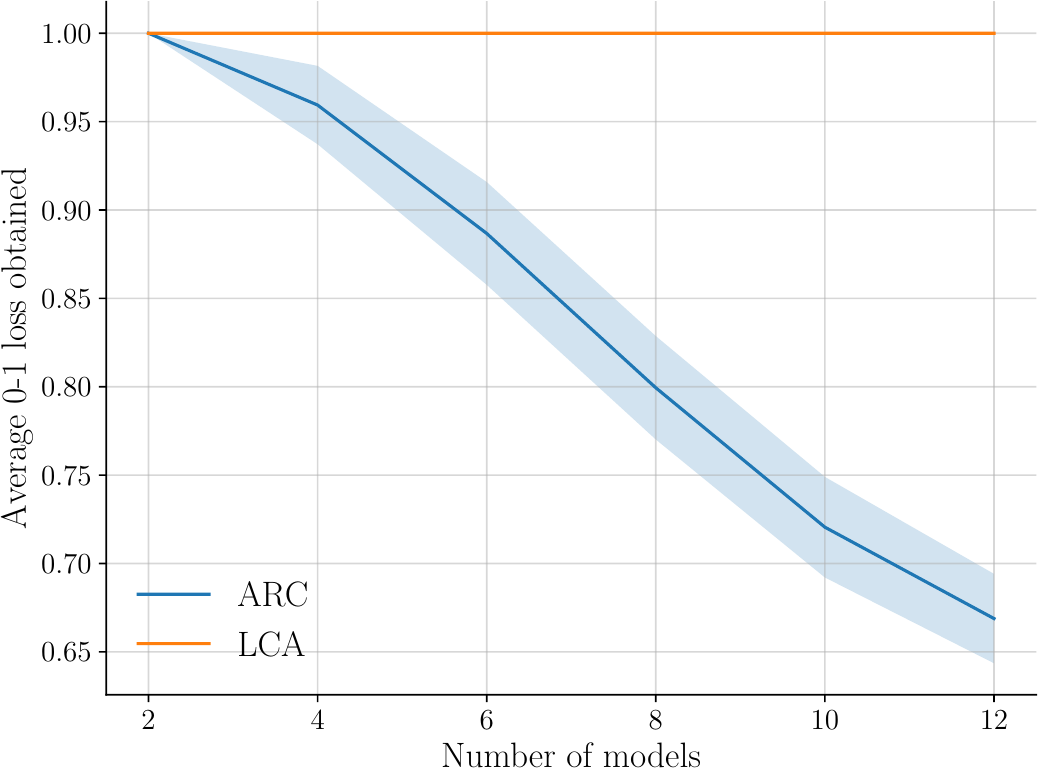}
    \caption{Biases sampled from $\cal N(0.2, 0.005^2)$. Classifiers are very close to the point and they are almost all at the same distance. In this particular case LCA performs very well, being always optimal. ARC on the other hand degrades as the number of models increases. The fact that LCA obtains a score of 1 indicates that all the models are always simultaneously vulnerable.}
\end{figure}

\begin{figure}[h]
    \centering
      \includegraphics[scale=0.5]{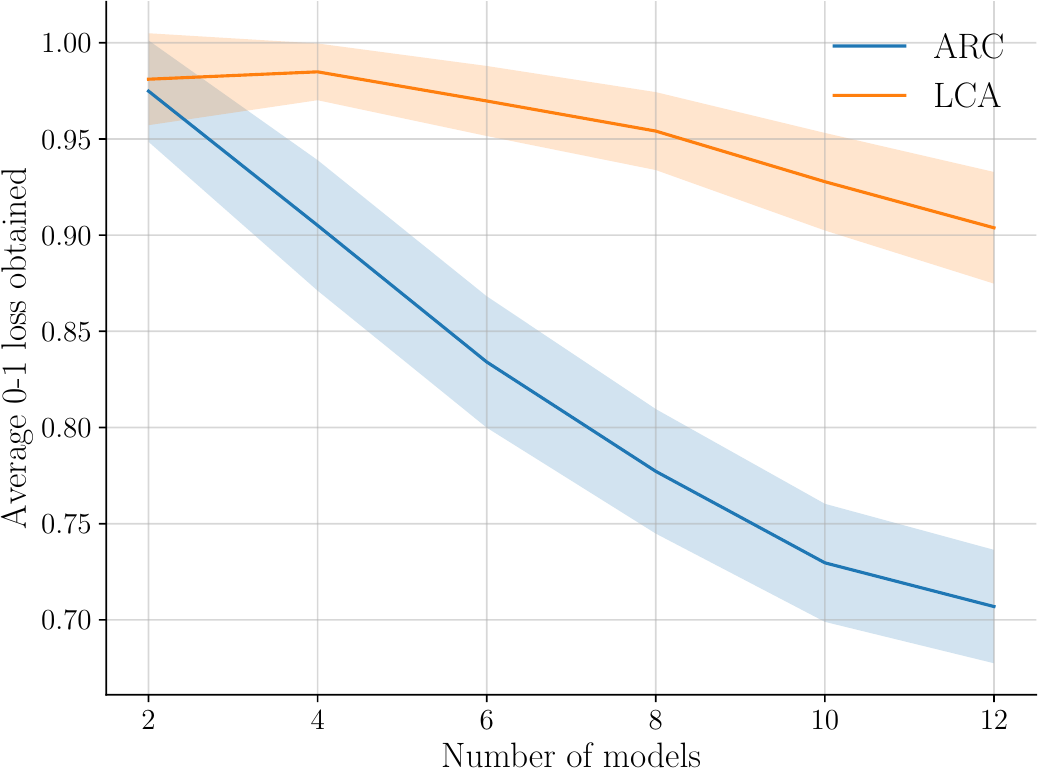}
    \caption{Biases sampled from $\cal N(0.2, 0.25^2)$. Classifiers are on average very close to the point but with a higher variance in their distance. LCA performs better, but the gap is less extreme than in the last case. In this scenario, LCA also degrades, but in what seems a slower rate than ARC.}
\end{figure}

\begin{figure}[h]
    \centering
      \includegraphics[scale=0.5]{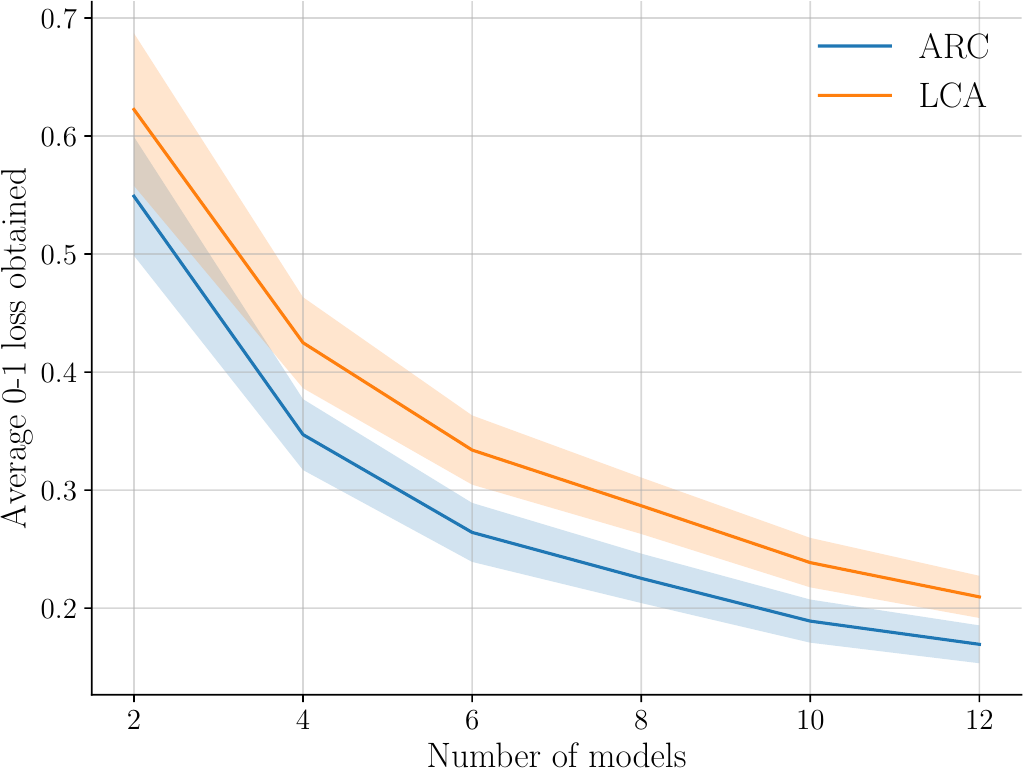}
    \caption{Biases sampled from $\cal N(0.8, 0.25^2)$. LCA is on average superior, but in this case the gap seems more constant than in other situations.}
\end{figure}

In terms of time, ARC in the binary linear case is very fast as it uses only $m$ gradient computations where $m$ is the number of classifiers. On the other hand, LCA performs $m \cdot T$ gradient computations, which makes it slower than ARC. Following the results on the convergence of PGD \cite{perdomosinger,bubeck2015convex}, a good choice for $T$ is in the order of $m^2$, in which case the complexity of LCA would be of order $m^3$.

\rule{\textwidth}{0.7pt}

\section{Training and attack parameters} \label{app:cifar}

The CIFAR-10 models (baselines and DVERGE) were taken from the original repository of DVERGE, available at https://github.com/zjysteven/DVERGE. We trained DVERGE(12) with 12 models using the implementation available in https://github.com/zjysteven/DVERGE and the following parameters:
\begin{itemize}
    \item \textbf{seed:} 0
    \item \textbf{batch size:} 512
    \item \textbf{epochs:} 200
    \item \textbf{learning rate:} 0.1
    \item \textbf{learning rate gamma:} 0.1
    \item \textbf{scheduler intervals:} [100, 150]
\end{itemize}

For the CIFAR-100 baseline models, the same parameters were used as for the CIFAR-10 baselines. For the DVERGE models, they were trained starting from the baselines using the default parameters. The parameter \textit{eps} used was 0.07 for the model with 3 and 5 classifiers, and 0.05 for the one with 8 classifiers. Batch sizes were 256, 1024 and 2048 for DVERGE with 3, 5 and 8 classifiers respectively, chosen because of resource availability. We did not optimize the batch size.

For GAL we also used the implementation available in https://github.com/zjysteven/DVERGE, and all default parameters were used except for the following: batch size was 1024, and the \textit{lambda} parameter was changed to 0.05 for GAL(3), and 0.2 for GAL(5) and GAL(12). This was done because the models with the default parameter were not achieving a good accuracy after 200 epochs.

For all the attacks we used the standard attacker budgets, \textit{i.e.} $\epsilon = 0.5$ for the $\ell_2$ threat model and $\epsilon = \frac{8}{255}$ for the $\ell_{\infty}$ threat model. All attacks have a $num_iters$ parameter that controls the number of gradient steps taken. It was set to $100$ for all three attacks. Particular parameters are:
\ \\ \\

\noindent
\textbf{APGD}
\begin{itemize}
    \item \textbf{eta:} 0.12549 for the $\ell_2$ model, 0.00784313 for the $\ell_{\infty}$ model.
    \item \textbf{num\_restarts:} 5
    \item \textbf{rand\_init:} True
    \item \textbf{momentum:} 0.9
    \item step size was multiplied by 0.5 at iteration $\lfloor 0.9 * \textit{num\_iters} \rfloor$.
\end{itemize}

\noindent
\textbf{ARC}
\begin{itemize}
    \item \textbf{step\_size:} 0.12549 for the $\ell_2$ model, $\frac{8}{255}$ for the $\ell_{\infty}$ model following \cite{dbouk2022arc}.
    \item \textbf{rand\_init:} False
\end{itemize}

\noindent
\textbf{LCA}
\begin{itemize}
    \item \textbf{eta:} 0.12549 for the $\ell_2$ model, 0.00784313 for the $\ell_{\infty}$ model.
    \item \textbf{num\_restarts:} 1 (no extra restarts)    
    \item \textbf{rand\_init:} False
    \item step size was multiplied by 0.5 at iteration $\lfloor 0.9 * \textit{num\_iters} \rfloor$.
\end{itemize}

\end{document}